\documentclass[10pt]{article}

\PassOptionsToPackage{breaklinks}{hyperref}
\PassOptionsToPackage{round, semicolon, authoryear}{natbib}
\PassOptionsToPackage{nameinlink}{cleveref}
\usepackage{hyperref}
\usepackage{arxiv}
\usepackage{macros}

\title{Audit, Alignment, and Optimization of LM-Powered Subroutines with Application to Public Comment Processing}
\author{%
Reilly Raab\\
Pacific Northwest National Laboratory\\
Richland, Washington, USA \\
\And
Mike Parker \\
Pacific Northwest National Laboratory\\
Richland, Washington, USA \\
\And
Dan Nally \\
Pacific Northwest National Laboratory\\
Richland, Washington, USA \\
\And
Sadie Montgomery \\
Pacific Northwest National Laboratory\\
Richland, Washington, USA \\
\And
Anastasia Bernat \\
Pacific Northwest National Laboratory\\
Richland, Washington, USA \\
\And
Sai Munikoti \\
Pacific Northwest National Laboratory\\
Richland, Washington, USA \\
\And
Sameera Horawalavithana \\
Pacific Northwest National Laboratory\\
Richland, Washington, USA \\
}

\usepackage[dvipsnames]{xcolor}

\usepackage{soul}
\newcommand{\both}[1]{\emph{\hl{#1}}}
\newcommand{\llm}[1]{\hl{#1}}
\newcommand{\sme}[1]{\emph{#1}}

\usepackage[most]{tcolorbox}
\usepackage[frozencache,cachedir=.]{minted}

\usepackage{booktabs}

\begin{document}

\maketitle

\begin{abstract}
The advent of language models (LMs) has the potential to dramatically accelerate tasks that may be cast to text-processing; however, real-world adoption is hindered by concerns regarding safety, explainability, and bias. 
How can we responsibly leverage LMs in a transparent, auditable manner---minimizing risk and allowing human experts to focus on informed decision-making rather than data-processing or prompt engineering?
In this work, we propose a framework for declaring statically typed, LM-powered subroutines (\ie, callable, function-like procedures) for use within conventional asynchronous code---such that sparse feedback from human experts is used to improve the performance of each subroutine online (\ie, during use).
In our implementation, all LM-produced artifacts (\ie, prompts, inputs, outputs, and data-dependencies) are recorded and exposed to audit on demand.
We package this framework as a library to support its adoption and continued development.
While this framework may be applicable across several real-world decision workflows (\eg, in healthcare and legal fields), we evaluate it in the context of public comment processing as mandated by the 1969 National Environmental Protection Act (NEPA):
Specifically, we use this framework to develop ``CommentNEPA,'' an application that compiles, organizes, and summarizes a corpus of public commentary submitted in response to a project requiring environmental review.
We quantitatively evaluate the application by comparing its outputs (when operating without human feedback) to historical ``ground-truth'' data as labelled by human annotators during the preparation of official environmental impact statements.
\end{abstract}

\section{Introduction}

Contemporary organizations have shown great interest in integrating language models (LMs) into workflows traditionally performed by human subject matter experts (SMEs), such as in medical diagnostics \citep{artsi2025large}, legal assistance \citep{padiu2024extent}, financial risk analysis \citep{llm-finance}, and governmental permitting or regulatory reviews \citep{phan2024rag}.
Despite this interest, however, the use of LMs (\eg, via a standard conversational interface) in high-stakes contexts is constrained by the need for decision-making reliability, objectivity, transparency, and accountability that SMEs currently provide \citep{mori2024large}.
Effective reconciliation between LMs and SMEs thus represents a critical frontier in real-world deployments of artificial intelligence.

LMs have demonstrated remarkable capabilities in extracting information from large volumes of multi-modal, multi-domain data; synthesizing multi-document concepts; and performing tasks associated with basic reasoning.
Nonetheless, LMs are susceptible to ``hallucinations'' (\ie, inaccurate generation) \citep{ji2023survey}, difficulty in handling nuanced, domain-specific requirements \citep{ashqar2025benchmarking}, historical biases inherited from training data \citep{ranjan2024comprehensive}, and opaque reasoning in decision-making \citep{machot2024building}.
Notably, these weaknesses are often precisely the strengths of SMEs, who are conversely burdened with the inefficient and labor-intensive tasks of cross-document, multi-modal search and information extraction.
We can see the need to delineate and integrate the often low-stakes or tedious work that can be performed by LMs with the discerning, high-stakes decision-making tasks performed by SMEs in the real world: The challenge is to harness the time efficiency and broad knowledge capabilities of LMs while preserving the domain expertise, contextual judgment, oversight, and accountability of SMEs.
Moreover, we must do so without creating additional burdens for SMEs to work with LMs (\eg, ``prompt-engineering'' or manual review of all LM tasks), and we wish to minimize the introduction of new risks (\eg, a loss of clarity regarding where or how LMs may be used by each SME, or, in the case of governmental work, the erosion of public trust).

In this work, we propose a novel auditable and interactive refinement framework for the effective integration of LMs with SMEs for decision-making workflows. 
In this framework, LMs are framed as execution engines within the well-scoped, text-based subroutines\footnote{In this work, our use of ``subroutines'' refers to named blocks of code that can be invoked with inputs of pre-specified type to produce outputs of pre-specified type, though the outputs may depend on internal state or randomness.}, which are auditable by SMEs. 
This formulation further accommodates sparse feedback provided by the SMEs to optimize the performance of individual subroutines, aligning work performed by the LM with aggregate SME expectations and preferences over time.

We demonstrate the applicability and effectiveness of this proposed framework in the context of the U.S. federal National Environmental Protection Act of 1969 (NEPA) permitting review process, particularly focusing on streamlining the public comment processing workflow. 
We selected the NEPA public comment processing as the test case due to its pressing need and significant social and environmental impact.
Currently, the public comment processing process is highly tedious and manual, resulting in substantial economic burdens and project delays for federal agencies. 
Our framework aims to address some of these challenges, thereby supporting the core objective of the NEPA (\ie, effective communication of the environmental impact of the proposed projects to benefit and inform the public). 
The need for AI integration in public comment processing is further reinforced by the growing interest of the U.S. federal government in streamlining federal permitting processes \citep{presidentmemo}.

\paragraph{\textbf{Primary Contributions}}
The contributions of this work are as follows:
\begin{enumerate}
\item A \textbf{methodology} for optimizing subroutine prompts combining a sampling method targeting infinitely many-armed bandit ($\infty$-MAB) regret minimization and self-critique loops to leverage sparse human feedback.
\item An open-source \textbf{software library}\footnote{A link to the library will be included pending internal disclosure processes.} that implements this contributed methodology, suitable for static-type analysis in Python and integrated with a relational database for tracking LM outputs and user feedback for online audit.
\item Initial \textbf{evaluation of an application}, called CommentNEPA, built with the contributed library, specialized for public comment processing.
\end{enumerate}

In this work, we develop a framework for the auditable use of LMs by SMEs, capable of dramatically accelerating the processing of public comments related to environmental permitting under NEPA. 
This technical framework, which we detail in \Cref{sec:technical}, benefits from sparse feedback provided by the SMEs to optimize the performance of individual subroutines, aligning work performed by the LM with aggregate SME expectations and preferences.
We compare the outputs achieved through the use of this framework, through an application called ``CommentNEPA,'' detailed in \Cref{sec:commentnepa}, to historical outputs achieved from SME labor (\Cref{sec:evals}).

\subsection{Application Domain: Comment Processing for Environmental Review}

Public participation in environmental review under NEPA plays a critical and statistically significant role in shaping decisions and outcomes in the United States \citep{Stava_2025}.
Unfortunately, comment processing for environmental review is tedious and labor-intensive for SMEs, resulting in delays that are a frequent target of political criticism.
Despite pressure for accelerated processing, demands on expert labor is further exacerbated by an increasing volume of projects and public comments, which now also leverage generative AI.
As we consider the use of generative AI to help SMEs accelerate comment processing while maintaining and supporting public engagement, we must acknowledge the sensitivity of this topic:
Inappropriate use of AI may introduce risks of litigation, a loss of public trust, or declining engagement, especially if concerns and comments are misconstrued or ignored.
Our emphasis on auditability and use of generative AI for text-processing and analysis---rather than autonomous decision making---is informed by this perspective.

Multiple agencies and organizations participating in NEPA-related work have systems to handle comments and responses. 
These range from simple spreadsheet-based workflows to specialized database tools (e.g., the U.S. Nuclear Regulatory Commission [NRC] uses a custom application for its comment and response work). 
NEPA practitioners have been quick to realize the potential of generative AI in this process and have been developing solutions with varying levels of success and maturity;
One example familiar to the authors include a comment-delineation tool created by Jacobs \citep{nally2025workshop}.

Among related work, \citet{april2024} sought to replicate human workflows with zero-shot static prompting of a fine-tuned LM, without using token sampling constraints to guarantee an output schema.
The authors concluded that zero-shot prompting to replicate human workflows faced limitations related to brittleness and inconsistency for this use case, but proposed in closing that an alternative might involve continuously improving, LM-powered steps with expert feedback.
The present work delivers on that proposal.
Beyond public comment analysis, our proposed framework and approach to risk-mitigation may be well-suited to further evaluations in specific medical or legal domains.

\subsection{Responsible Use of Language Models}

Two competing paradigms have emerged in the race to leverage generative LMs for tasks with economic value. 
The first paradigm views LMs as execution engines for well-scoped, text-based subroutines within the context of traditionally managed state and logic (e.g., by invoking an LM via an API call from a traditionally executed program, optionally restricting the LM return with a formal grammar\footnote{A capability \citep{structred-generation} referred to as ``structured generation,'' ``constrained generation,'' ``guided generation,'' or ``JSON-mode.''}).
The second paradigm, envisioning LMs as a path towards general, autonomous ``agents,'' treats the auto-regressive output tokens of the LM as an action space capable of autonomously invoking so-called ``tools'' or effecting state-changes that are not managed by traditional code and may not be open to review or facile interpretation.
Such an approach attempts to identify LM auto-regressive generation with cognitive models of ``thinking'' or ``reasoning'' in the completion of goals with wide scope and open-ended structure.

We strongly prefer the first paradigm when using LMs for tasks with potentially non-trivial consequences and thus develop our work on fine-grained subroutine optimization and auditability.
That is, to mitigate risk, we explicitly adopt an approach that minimizes the autonomous ``decision-making'' that LMs perform.
Instead, business logic and control flow are implemented in traditional application code, and LMs are treated as execution engines for text-based subroutines.
To minimize SME labor, we focus on improving the performance of these individual subroutines through SME feedback during an online review and audit process.

To effectively use LMs as execution engines for text-based subroutines in practice, we start with two observations about contemporary models:
First, the commercialization of LMs and suitable APIs for executing individual functions has converged toward models that are fine-tuned for ``instruction following,'' whereby the output of the model is conditioned on instructions provided by a ``system prompt'' that is separately delimited from the rest of the initial context window.
In general, instruction-tuned LMs are highly sensitive to the system prompt they are provided \mbox{\citep{promptbreeder, vertexai}}.
Rather than reinvent interface for LM-powered subroutines, we adopt this ``system-prompt'' model out of convenience and practicality.
Second, fine-tuning the (billions of) weights that parameterize the most capable LMs for specific tasks is a costly endeavor, even using techniques such as low-rank adaptation \citep{lora}.
Given that we wish to perform a multiplicity of subroutines with LMs, we avoid fine-tuning our own model weights and choose instead to optimize the system prompt used for each subroutine.

\subsection{Related Work}

We briefly review research in LM-powered subroutine optimization as it relates to the present work. 
To our knowledge, our treatment of prompt selection as a many-armed bandit (MAB) problem over infinite arms is novel. 
Nonetheless, we do not identify this as our primary contribution (the method we adopt for this work is simplistic and easy to reason about during development); rather, our focus is on building auditable applications, incorporating LM-powered subroutines, that align their performance with online human feedback.

\paragraph{Prompt Search Methods}
The space of research for generating prompts for LM subroutines has seen rapid development in the last three years.
While methods from machine learning may be used to optimize over continuously parameterized (so-called ``soft'') prompts \citep{softprompts}, such techniques are typically incompatible with LMs served over API result in uninterpretable text \citep{protegi}, limiting the capacity for human oversight and audit.
Research in prompt optimization has therefore focused on gradient-free optimization methods over discrete sets.

Early efforts at gradient-free prompt optimization demonstrated that LMs themselves can be leveraged to generate, select, and mutate effective prompts \citep{ape}.
Subsequent research has proposed a host of optimization techniques for searching the space of prompts, handling feedback signals, incorporating LMs in the prompt optimization framework, and optimizing over programs that comprise multiple LM subroutines.
Some methods treats the space of prompts as combinatorial (\eg, with respect to independently optimized ``instructions'' and ``examples'') \citep{vertexai}, while others construct prompts through tree-search \citep{promptbreeder} or beam-search \citep{protegi} methods over mutation operations \citep{promptbreeder} or phrases \citep{grips}, optionally incorporating elements of planning \citep{promptagent}, synthetic data \citep{sipdo}, or semantic embedding distances \citep{prompt-stability}.

In contrast to such sophisticated search methods, our work takes the simple approach of identifying the best candidate from the set of prompts that a given LM may stochastically generate (\ie with sufficiently high ``temperature'') conditioned on a static description of the LM-powered subroutine that we wish to optimize (\Cref{sec:technical}).
While this approach does not enforce specific prompt structure, such as explicit instructions followed by examples that can be independently optimized (as in \citet{vertexai}), it is not clear to what extent such a constrained structure is necessary, nor how much it is future proof.
For example, without being prompted to do so, our LM-based prompt generation regularly synthesizes examples.
We claim that this behavior is unsurprising if we assume that effective LM prompt techniques are included in the corpus used the train the underlying model, which becomes increasingly likely with time.

\paragraph{Online Prompt  Optimization and Many-Armed Bandits}
For our use case, we have assumed that evaluating prompt performance ultimately requires the expert human labor, which we seek to use efficiently. 
Therefore, we wish to optimize prompts \emph{online}, unifying SME audit and review with feedback used to further improve our LM-powered subroutines.
While beam-search or Monte Carlo tree-search methods are compatible with online optimization, such methods must bias new prompts to be ``closer'' to the most performant past examples.
Without structuring prompts to adhere to specific templates or phrase combinations, such an approach requires using an LM to generate new prompts given examples of past prompts, as achieved by \citet{promptbreeder, protegi, promptagent}.

Despite our attempts to initially pursue a similar strategy, we found that the ``meta-reasoning'' capabilities of current LMs were then insufficient to avoid confusion when mutating prompts of the type we wished to use in our application (\ie, with non-trivial type constraints and domain-specific context).
That is, when we tasked an LM with generating a novel prompt for a realistic target task, such as extracting a JSON-structured list of concerns with supporting quotes from a letter of public correspondence, given in-context prior examples of prompts and their associated feedback or ratings, the LM often repeated the top prompt without meaningful variation (thus losing diversity) or else incorporating elements of its own prompt, especially with respect to the descriptions of desired types.
Our contributions are therefore based on a MAB formulation of the prompt-selection problem.

\citet{protegi, triple} have previously utilized MAB (specifically, best-arm identification) formulations for prompt optimization:
\citet{protegi} combine an upper-confidence bound (UCB) sampling method over finitely many prompts to guide beam-search, refining sampled prompts with textual gradients.
\citet{triple} use a MAB formulation for best-arm identification with fixed budget, but we do not assume fixed budgets and have potentially infinite number of prompts.
Our work is distinguished by a formulation that allows for a variable (growing) number of potentially-infinite ``arms'' (\Cref{sec:technical}).
Moreover, we adopt the regret-minimization objective rather than best-arm identification, given that regret corresponds to work that SMEs must perform to correct LM outputs.

\paragraph{Self-Critique and Feedback Propagation}
While the MAB problem has a scalar-valued objective, we recognize that natural language can provide much richer information than scalar feedback signals if SME feedback on LM outputs could be appropriately used to guide the generation of new prompts.
By analogy to back propagation, \citet{textgrad} pioneer the use of an LM to generate natural-language feedback on the inputs or prompts of a subroutine given natural-language feedback on its output.
While we hope to incorporate of such mechanisms (along with beam-search) into future revisions of our framework, our approach to propagating scalar-valued feedback on outputs to prompts is detailed in \Cref{sec:technical} and is structured to accommodate ``self-critique loops'' that are structured to align LM-generated feedback on intermediate outputs with SME preferences.

The use of a self-critiquing or self-refining LM is not novel: \citet{self-refine} propose an iterative process where an LM generates an output, then critiques it, and subsequently refines it based on this self-generated feedback. 
Nonetheless, we believe our work is novel in optimizing the critique task to specifically align its feedback with sparse human judgments.
This innovation may be considered an application of reward modeling, analogous to reinforcement learning with human feedback (RLHF) \citep{rlhf}:
One subroutine is optimized to learn a model of human feedback, and this subroutine is used to generate feedback to optimize another that actually performs the target task.

\paragraph{Optimizing LM Programs with Structured Outputs}
Our emphasis on LMs as execution engines for well-scoped subroutines often necessitates structured or constrained outputs. Techniques for guiding LM generation using formal grammars (e.g., JSON schema, BNF grammars) \citep{structred-generation} are now widely supported by commercial APIs.
While frameworks like DSPy \citep{dspy} have emerged to systematize the development of LM-based applications by creating pipelines of LM "modules" (prompted calls) and optimizing their components, including prompts and few-shot demonstrations. While DSPy provides modular approach to optimizing entire LM programs, individual prompts, and model weights, its current mechanisms do not support hard constraints that integrate with static type-checking at time of writing.

To summarize the relationship of our work to prior literature, our contributions distinctively combine online, MAB-based prompt selection for individual subroutines, concurrent optimization of base and human-aligned self-critique tasks, and a data-dependency tracing mechanism for auditability and feedback propagation. This approach is tailored for the paradigm of LMs as reliable, auditable execution engines for fine-grained, text-based subroutines, particularly where sparse feedback is the primary signal for improvement.

\section{Methodology}
\label{sec:technical}

We highlight four technical contributions that comprise our use of LM-powered subroutines, individually optimized with sparse human feedback during online audit:
\emph{First}, we treat the selection of system prompts for each subroutine as a MAB (online regret-minimization) problem over a potentially infinite number of potential prompts (\Cref{sec:prompt-sampling}).
\emph{Second}, we use an LM to generate novel prompts, on demand, for subroutine declared by normal code (in Python) with type guarantees recognized by static type-checking (\Cref{sec:subroutine-declaration}).
\emph{Third}, we identify two distinct types of subroutines that we optimize concurrently: a base task and a self-critique task that provides feedback on an associated base task. The self-critique task is optimized to agree with sparse human feedback (\Cref{sec:self-critique}).
\emph{Fourth}, we propagate feedback on LM outputs to the generating prompts by tracing automatically recorded, mutual data-dependencies between LM outputs that enable interpretability (and audit on-demand \Cref{sec:audit}). 
A high-level description of how we implement these contributions in practice is described in \cref{sec:library}.

\subsection{Online Prompt Sampling}
\label{sec:prompt-sampling}

We formalize the problem of sampling prompts for each subroutine as an $\infty$-MAB problem.
This problem is formalized in terms of minimizing cumulative loss (regret) over a (possibly infinite) number of time-steps, where, at each time-step, we are given the choice to ``pull'' one of many ``arms'' (so-called because the motivating picture is of a row of slot-machines, where pulling an arm gives a randomly sampled award and each arm may have a different distribution from which rewards identically and independently sampled):
Solutions to this problem must balance the need to explore (to find the best arm) with the need to exploit (use the best arms found so far).

It is standard, when dealing with a finite number of arms, to pull each possible arm once to initialize estimates for the expected loss for each arm before proceeding with a sampling-based approach such as upper confidence bound (UCB), Thompson Sampling, or Boltzmann exploration (BE).
When dealing with an infinite number of arms however, this is infeasible, and so we adopt the strategy of combining all ``unexplored'' arms into a single, substitute \emph{exploration} arm:
When the exploration arm is pulled, a new arm (uniformly sampled from the set of unexplored arms) is selected and pulled instead, while the \emph{exploration} arm persists at all time-steps.
At each time step, we need apply our sampling algorithm only to the set of previously explored arms augmented by the exploration arm.

In practice, we opt for the simplicity of Boltzmann exploration (also known as ``Softmax sampling,'' which forms the basis of EXP3 \citep{exp3}) with an (inverse) temperature \(\beta\).
At each time step, we estimate the loss of the exploration arm (that is, generating and using a novel prompt) by the population average loss of all arms pulled thus far. 
Symbolically, for a set of previously explored arms \(\{x_1, x_2, ..., x_n\}\) and an ``explore-arm'' \(x_0\), the probability of sampling \(x_i\) is
\begin{equation}
\label{eq:sampling}
\forall i \in \{0, 1, ..., n\}, \quad \Pr(x_i) = \frac{
\displaystyle{\exp}\big(-\beta \scrL_i)\big)
}{
\displaystyle
\sum_{j=0}^{n} {\exp}\big(-\beta \scrL_j\big)
}; \qquad
\scrL_{0} = \frac{1}{n} \sum_{j=1}^n \scrL_j,
\end{equation}
where \(\scrL_i\), for \(i \in \{1, 2, ..., n\}\), represents the average loss for arm \(x_i\) and
\(\scrL_{0}\) is the estimated loss of the ``exploration'' arm.
We wish to minimize the cumulative value of
\begin{equation}
\sum_{i=0}^n \Pr(x_i) \scrL_i,
\end{equation}
which is the same objective as minimizing the cumulative regret.
Unfortunately, it is known that the performance of sampling algorithms for bounding this value depend significantly on the actual (or assumed) distribution of losses \citep{infinite-mab-simple}, about which we have little knowledge for our use-case.

Nonetheless, we can empirically show how distribution of prompts evolves in time:

\paragraph{A Demonstration of Prompt Evolution}
As an example of how the sampling routine provided in \Cref{eq:sampling} results in an evolution distribution of prompts for a subroutine, let us declare a subroutine that counts the number of words in a sentence that contain ``rare letters,'' but without declaring which letters are rare (\Cref{fig:example-subroutine}). 
In this way, we approximate a subjective task which may nonetheless be compared against ``ground-truth'' (a correct, deterministic answer provided by fixing ``rare letters'' as the set \(\{\text{Q}, \text{W}, \text{X}, \text{Z}\}\)) to drive updates to the underlying distribution of prompts.

\begin{figure}[h!]
\centering
\includegraphics[width=4.5in, trim={0 0 0 2em},clip]{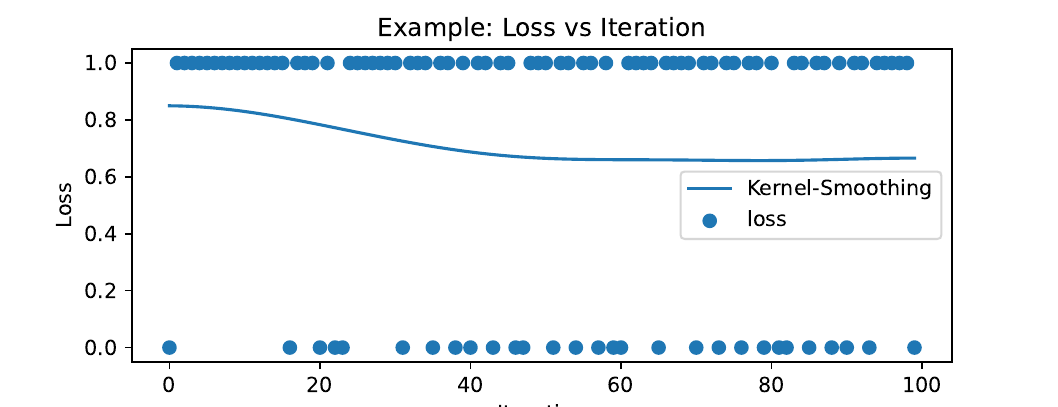}
\caption{We track a decrease in average loss over the course of 100 trials for an LM-powered subroutine with system prompt selected according to \Cref{eq:sampling} at each trial. Smoothing of the discrete loss values is achieved with a Gaussian kernel with \(\sigma = 15\). As expected, the average loss decreases over time.}
\label{fig:loss-vs_iteration}
\end{figure}

\begin{figure}[h!]
\centering
\includegraphics[width=4.5in,trim={0 0 0 2em},clip]{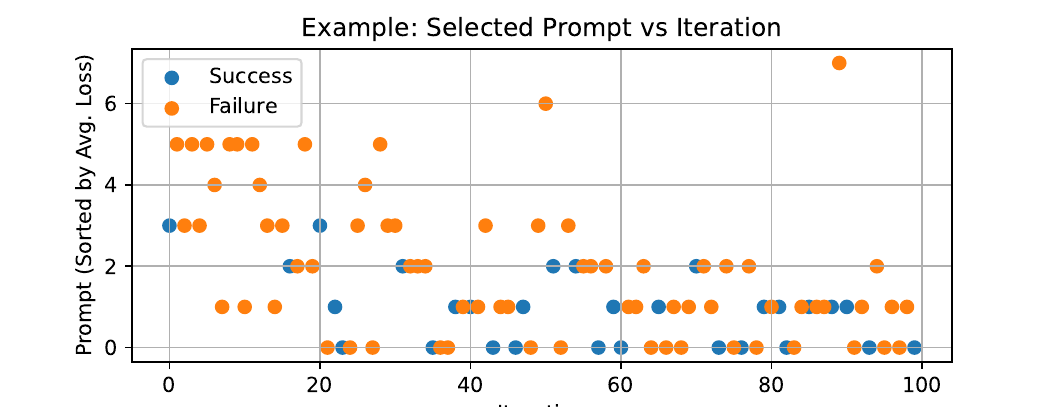}
\caption{We track which prompt (sorted vertically by empirical average loss) is selected for each trial of a target subroutine. As expected, the prompts with lower average loss are selected more frequently over time. The prompt with the lowest average loss is shown in \Cref{fig:example-prompt}.}
\label{fig:prompt-vs-iteration}
\end{figure}

In our experiment, we assign a loss of 1.0 to examples that provide the incorrect word count relative to the ``ground truth'' answer) and linearly increase the value of \(\beta\) from 0 to 1 over the course of 100 trials while sampling prompts according to \Cref{eq:sampling}.
We visualize the results.
\Cref{fig:loss-vs_iteration} tracks the loss achieved over sequence of trials (randomly selecting from a set of sentences that contain at least one instance of each letter, as in ``Pack my box with five dozen liquor jugs.''), while \Cref{fig:prompt-vs-iteration} tracks which prompt was selected for use during each trial.
The prompt with the lowest average loss is shown in \Cref{fig:example-prompt}.

We leave further refinements, such as more sophisticated scheduling of \(\beta\) or a replacement of Boltzmann sampling with Maillard sampling (known to improve regret bounds over Boltzmann sampling in the finite arm case with sub-Gaussian reward distributions; \citet{maillard}), to further development of this work.

\subsection{Lazy Prompt Generation from Subroutine Declaration}
\label{sec:subroutine-declaration}

We declare a subroutine by combining an input type specification, output type specification, and documentation string, using an interface intended to replicate function declaration in traditional code.
Concretely, we implement our framework in Python, using Pydantic \citep{pydantic} to define the input and output type specifications (also known as ``schemas''):
We provide an example in \Cref{fig:example-subroutine}:
\begin{figure}[h!]
\begin{tcolorbox}[title=Example Subroutine Declaration]
\begin{minted}{python}
class Input(pydantic.BaseModel):
    given_text: str = Doc("The given text")

class Output(pydantic.BaseModel):
    scratch_work: str = Doc("A place for scratch work")
    character_count: int = Doc("Number of instances in the text")

class MyAgent(framework.Agent[Input, Output]):
    """
    Count the of number of words containing rare letters in the given text.
    """
\end{minted}
\end{tcolorbox}
\caption{An example subroutine declaration in our framework. This declaration is used to synthesize prompts, such as the one documented in \Cref{fig:example-prompt}, by querying an LM.}
\label{fig:example-subroutine}
\end{figure}

When a subroutine is invoked, an LM taking on the role of ``prompt engineer'' synthesizes candidate prompts for target subroutines by using a hard-coded system prompt of its own and accepting a subroutine declaration and additional optional ``context'' as input. 
Following the example of \Cref{fig:example-subroutine}, the prompt in \Cref{fig:example-prompt} was synthesized by an invocation of the LM ``prompt engineer.''

\begin{figure}[h!]
\begin{tcolorbox}[title=Example LM-Generated Prompt]
\begin{minted}{markdown}
You are an expert text analyst. Your task is to analyze a given text and identify the 
number of words that contain rare letters.

Here's how you should approach the task:

1.  **Receive the Input:** You will be given a text string as input.

2.  **Identify Rare Letters:** Define "rare letters" as any letters that are not 
    commonly used in the English language. (e.g., J, X, Q, Z)

3.  **Word Analysis:** Examine each word in the text to determine if it contains any
    of the rare letters.

4.  **Count Words with Rare Letters:** Count the number of words that contain rare 
     letters.

5.  **Output the result in JSON format:**  Your response must be a JSON object 
    adhering to the following schema:

    ```json
    {
      "scratch_work": "...", // A place for your reasoning and analysis.
      "character_count": integer // The final count of words with rare letters.
    }
    ```
\end{minted}
\end{tcolorbox}
\caption{A prompt synthesized by an LM for the subroutine declared in \Cref{fig:example-subroutine}. 
In practice, such LM-generated prompt often contain synthetic examples, despite the fact that we neither direct nor constrain such behavior:
We presume that contemporary LMs are now trained on examples of various prompting techniques.
This prompt minimized loss on our example task by correctly identifying the letters we secretly defined as ``rare'' (J, X, Q, Z), serving as a proxy for the ability stochastically uncover apriori-unknown preferences of SMEs.
We provide the code used to serialize the declaration and the prompt used for the LM ``prompt engineer'' in our release of an open-source library (\Cref{sec:library}).}
\label{fig:example-prompt}
\end{figure}
In addition to the (purely LM-generated) prompt, the required output type for the declared LM subroutine is forwarded to the API to constrain the return with a guarantee that it will adhere to the correct schema. 
In general, we implement our framework such that it works with static type-checking tools such as pyright \citep{pyright}.

\subsection{Optimization with Self-Critique and Sparse Human Feedback}
\label{sec:self-critique}

In building the capability to sample prompts based on their expected loss, we have assumed that loss is something that we can easily measure or infer.
When we do not have objective measure of subroutine performance, we would like to leverage human expertise but not overburden SMEs with the need to review many LM outputs for each prompt to provide a robust signal.
Our strategy is to define a separate subroutine that ``critiques'' the performance of another, and align the performance of the critique with human preferences that may be ``sparse'' (in that we would like only a few human ratings to be leveraged to provide many more ratings on the subroutine that is being critiqued).
The possibility of defining a self-critique subroutine that can be used for numerical feedback is enabled by constrained generation.
We provide an example output type to be used in the declaration of such a subtask in \Cref{fig:example-critique-rating}.

\begin{figure}[h!]
\begin{tcolorbox}[title=Example Critique Output Type]
\begin{minted}{python}
class CritiqueOutput(pydantic.BaseModel):
    explanation: str = framework.Doc("Justification for the provided rating(s)")
    brevity_rating: IntEnum = framework.Doc("rating for brevity (int from 1 to 10)")
    ... # type may be extended to allow ratings along multiple desiderata
\end{minted}
\end{tcolorbox}
\caption{An example output type used to constrain an LM-powered ``critique'' subroutine. The
associated input type consists of an input-output pair for the target subroutine receiving the critique. By generating explanations together with ratings, we attempt to provide interpretability for the SME's audit and leave open the possibility of refining prompts with ``textual gradients'' \citep{textgrad} in extensions of the present work.}
\label{fig:example-critique-rating}
\end{figure}

\begin{figure}[h!]
\centering
\includegraphics[width=4.5in]{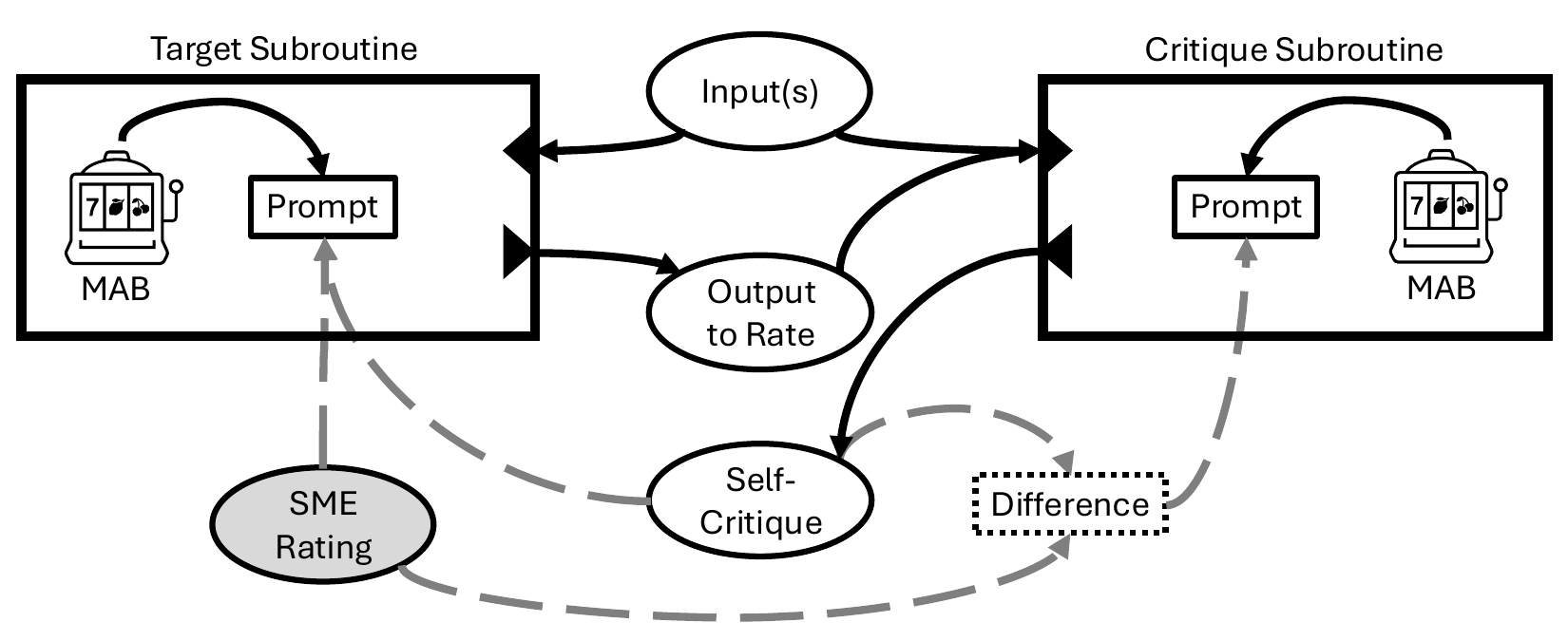}
\caption{Outputs from a ``critique'' LM-powered subroutine are used to adjust \(\scrL\) of the prompt for another LM-powered subroutine by evaluating the output it generated. The prompts for the critique subroutine are evaluated on the basis of minimizing disagreement with an SME performing the same task of evaluating the outputs of the target subroutine.}
\label{fig:base-vs-critique}
\end{figure}

We achieve this by using each SME rating of an LM-generated output to derive a loss signal for a LM-generated self-critique of the same output, visualized in \Cref{fig:base-vs-critique}. 
Specifically, we infer and apply the loss value
\begin{equation}
    \scrL_{\text{critique}} = \Big(\scrL_{\text{target}}[\text{SME}] - \scrL_{\text{target}}[\text{critique}]\Big)^2,
\end{equation}
where the values in square brackets indicate where the loss signal originates.
To minimize this loss, self-critiques must accurately align with expected SME evaluations, while these self-critiques are further used to optimize upstream subroutines.
This combination is the central technique we use to optimize LM-powered subroutines with sparse human-feedback, and which we compare to reward-modeling techniques like reinforcement learning from human feedback (RLHF; \citet{rlhf}).

\begin{figure}[h!]
\centering
\includegraphics[width=4.5in,trim={0 6em 0 7em},clip]{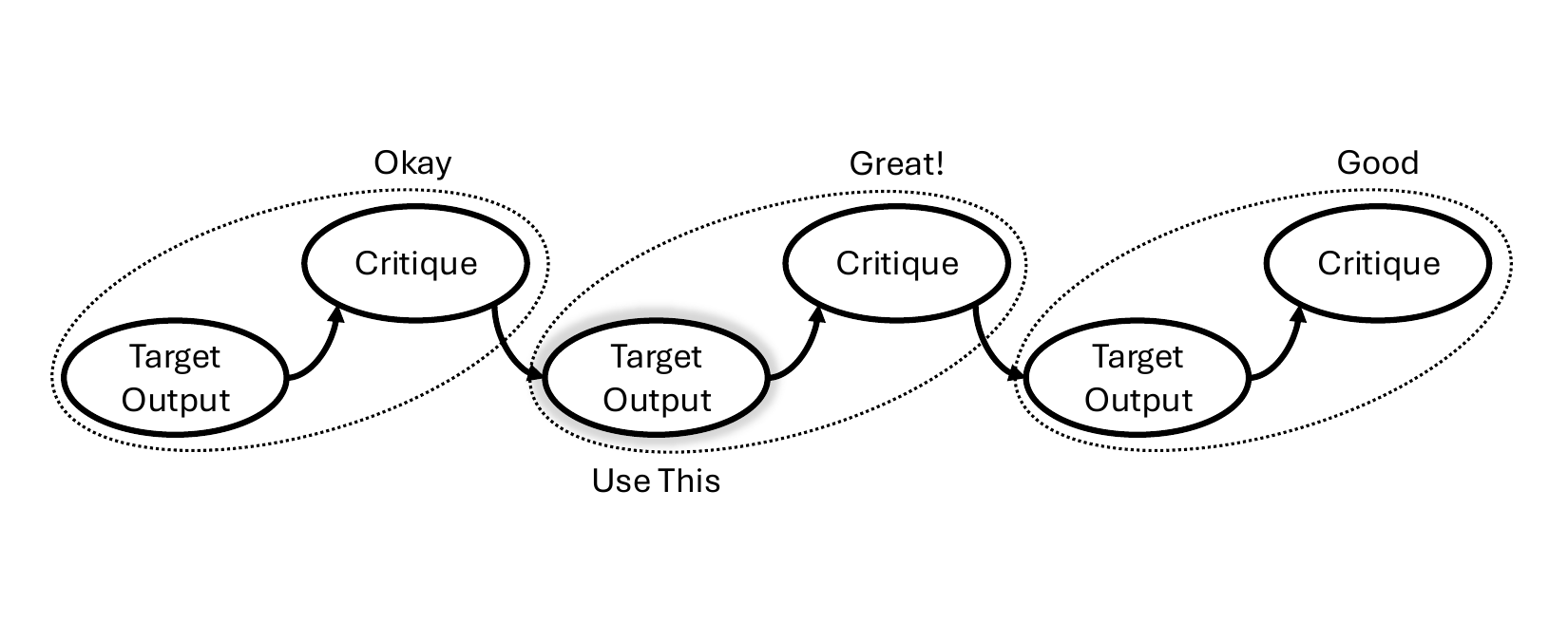}
\caption{In addition to optimizing the prompts used to perform each LM-powered subroutine, we optimize the outputs for a given task by iterating a ``self-critique loop.'' This multiplies the amount of feedback driven by self-critique. The output with the best self-critique evaluation is used for later processing stages. In practice, we guarantee a finite number of loop iterations by bounding and discretizing critique values (\Cref{fig:example-critique-rating}), and we allow for early exit of the loop when rating stop improving or are sufficiently high.}
\label{fig:self-critique-loop}
\end{figure}

In practice, we extend self-critique over multiple iterations in self-critique loop that both multiplies the amount of feedback the system generates for itself and allows us to select the best revision among competitors from a single task. 
This self-critique loop is represented by \Cref{fig:self-critique-loop}. 
We pass the result that achieves the best self-critique rating to subsequent processing steps in the application pipeline.

\subsection{Online Auditability}
\label{sec:audit}

Our library automatically records all inputs, prompts, outputs, and feedback with a relational database as LM-powered subroutines are invoked or used to provide mutual feedback each other. 
In particular, by wrapping subroutine calls in a data structure that includes pointers to the relevant database entries and allowing LM-powered subroutines to accept these wrapping data structures as arguments, data dependencies between LM-powered subroutine call are recorded, building up a computational graph that can be traced to expose the sequence of intermediate steps in the processing pipeline and any ``explanations'' or ``justifications'' generated by subroutines along the way.
We document this extensible database schema with our release of the library.

\begin{figure}[h!]
\centering
\includegraphics[width=4.5in]{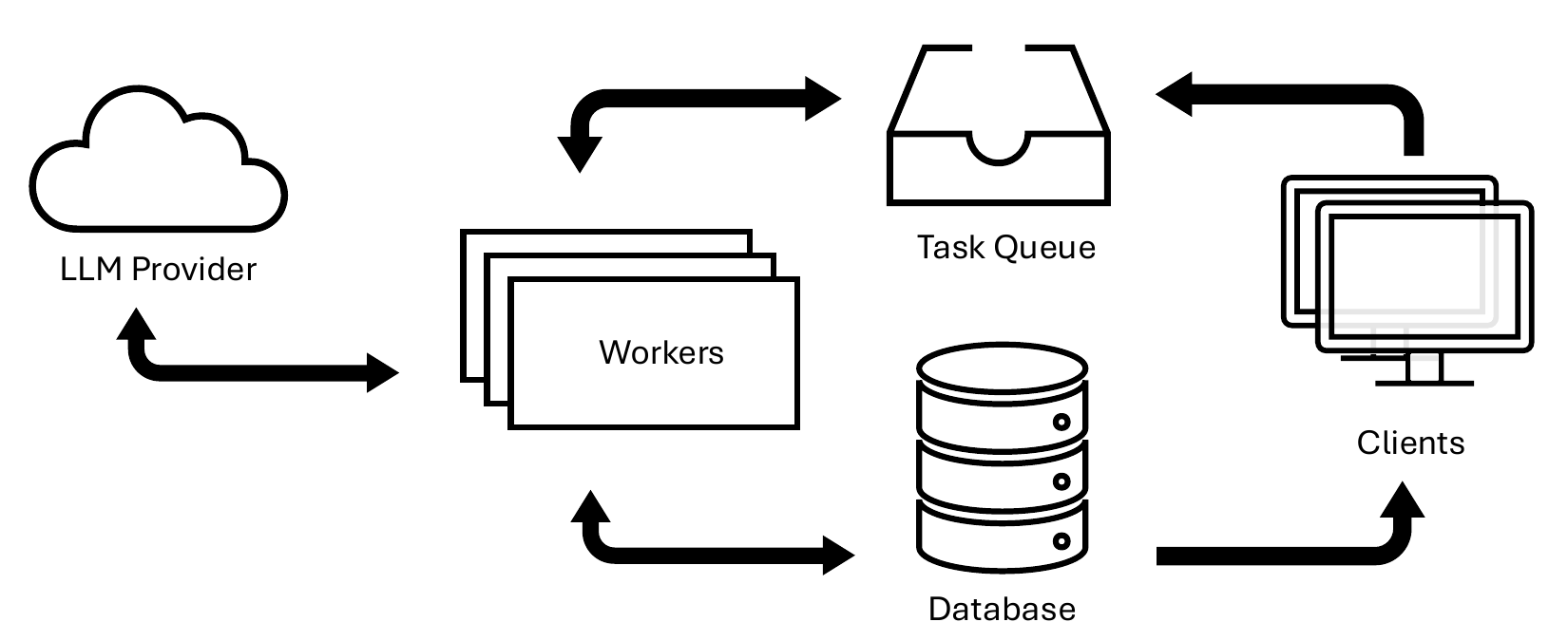}
\caption{A high-level overview of the architecture we implement to build applications on top of our proposed methodology.
We use a single database as a source-of-truth for shared state across multiple concurrent clients. 
Tasks are serialized to a durable message queue and executed concurrently by a pool of workers with the ability to call LM-powered subroutines, transact with the database, and spawn additional tasks.
We note that future development may benefit from the use of a so-called ``workflow engine'' \citep{van2003workflow}, which enables durable orchestration of distributed tasks and supports complex control flows beyond simple queuing mechanisms.
}
\label{fig:architecture-overview}
\end{figure}

\subsection{Open-Source Implementation and Architecture}
\label{sec:library}

A few practical considerations inform how we concretely implement the proposed methodology to build applications: 
First, we expect our applications to be input-output constrained (given the processing latency of individual LM requests); 
to allow concurrent processing, we demand an asynchronous interface for all LM-powered subroutines. 
Second, anticipating errors, we use a durable message queue for managing asynchronous jobs so that we can recover state if the process executing the LM-powered subroutine halts. 
Third, we want a durable, universal source of truth for the state of the application, for which we use a relational database.
Finally, while other languages may be more natural fits for this architecture, we wish to easily interface data-analysis and machine-learning tools, for which Python is the de-facto standard language.
We visualize how the concrete components in \Cref{fig:architecture-overview}.

\section{CommentNEPA}
\label{sec:commentnepa}

We use the methodology outlined in \Cref{sec:technical} to develop ``CommentNEPA,'' an application intended to accelerate the processing of public comments submitted in response to energy-related proposals during federally mandated environmental reviews.

\textbf{Pipeline} CommentNEPA breaks the comment-processing pipeline into four stages, delivering outputs to SMEs that assist the generation of a ``comment-response reports'' (documents that summarize, detail, and respond to the range of substantive commentary received from the public during a mandated review period).
\begin{enumerate}
    \item Summarize: Each letter is summarized, distilling its core concerns.
    \item Extract Concerns and Quotes: Concerns and supporting quotes for each letter are extracted.
    \item Bin Concerns: Each concern is mapped to one or more ``bins'' of similar concerns.
    \item Summarize Bins: Each bin is summarized with citations to the original letters.
\end{enumerate}
Each stage of this pipeline leverage uses our proposed framework for optimizing LM-powered subroutines with sparse human feedback during \emph{online} review / audit that proceeds in batches.

\begin{figure}[h!]
\centering
\includegraphics[width=4.5in,trim={0 2em 0 4em},clip]{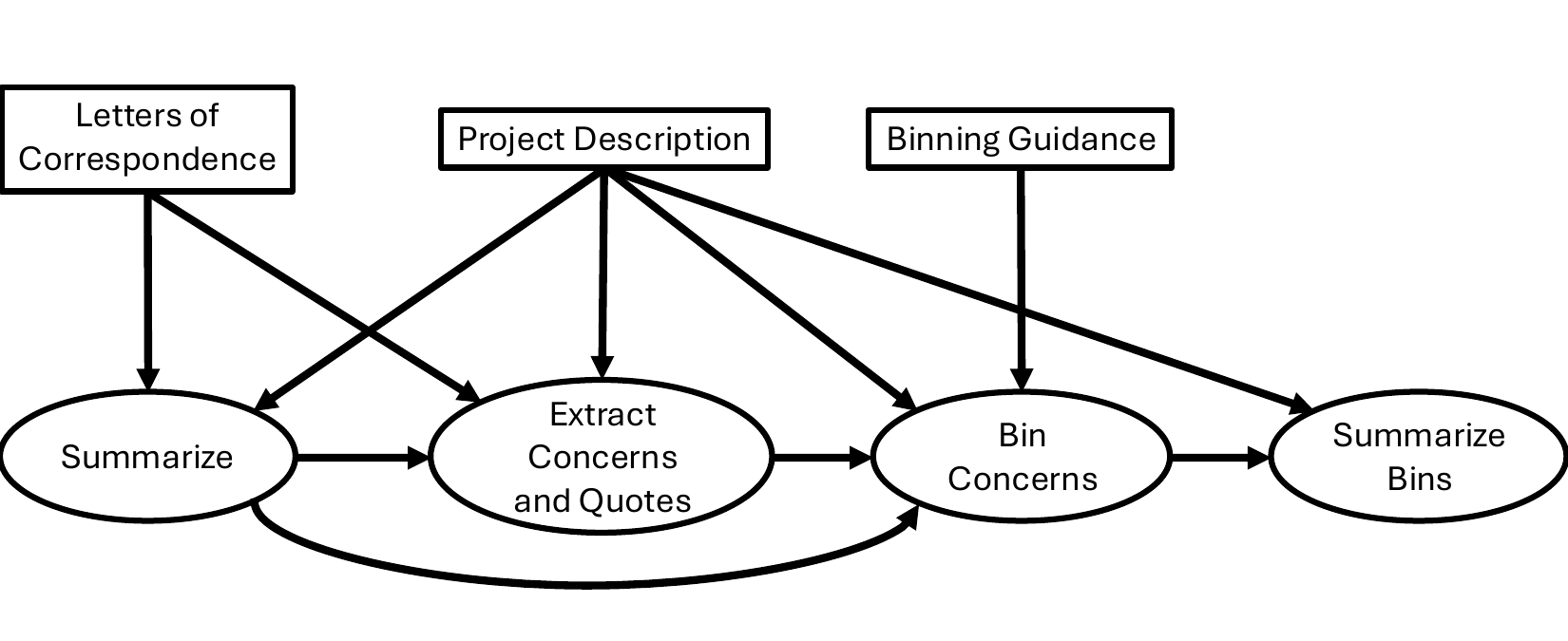}
\caption{Data flow between stages of the processing pipeline that we formalize for CommentNEPA. We treat the corpus of public correspondence, a description to which this correspondence pertains, and ``binning guidance'' requested by the SME users as inputs to the application.
Each stage implements a self-critique loop that may be optimized by sparse human feedback (\Cref{sec:self-critique}).}
\label{fig:commentnepa-pipeline}
\end{figure}

While our application is built to accommodate online audit and review at any stage, such that feedback affects prompt distributions as soon as it is received, the interface we initially provide is constructed around batched processing, whereby a team of SMEs can collectively review a batch of data and provide feedback before processing the next batch with the revised prompt distribution (incorporating feedback from the previous batch).
For CommentNEPA, this means that after each batch (\eg, of 100 letters), SMEs have the opportunity to review and provide feedback on the generated summaries, extracted concerns, selected quotes, or mapped bins to provide feedback before processing the next batch.
Feedback thus influences the prompt distributions for relevant subroutines when processing the next batch of letters.

\paragraph{1. Summarize}
For each letter of correspondence, the application first summarizes (in a self-improving evaluation-feedback loop) the letter to distill its chief concerns without regard to authorship, tone, or extraneous information.
This is intended as a denoising step to normalize public commentary and provide the necessary context to condition downstream tasks.

\paragraph{2. Extract Concerns and Quotes}
After summarizing each letter, the application uses this summary together with the original letter to extract a list of ``concerns,'' articulated in natural language, each of which is supported by a list of quotes from the original letter.
We post-process each quotation using fuzzy string matching \citep{rapidfuzz} to guarantee that it appears in the source letter.

\paragraph{3. Bin Concerns}
One of the primary objectives of a comment-response report is to identify the breadth and internal nuances of semantically distinct categories of public concerns.
In practice, these categories are referred to as ``bins'' and given names like ``terrestrial ecology'' or ``cultural resources.''
SMEs must mutually agree on the appropriate bounds of these mutually exclusive categories to effectively collaborate on the comment processing, and the standard practice is to document ``binning guidance'' that provides directions or examples to SMEs to ensure the compatibility of their independent work.

Because this guidance takes the form of natural language instructions, CommentNEPA is designed to accept this guidance as an input.
Specifically, it is part of the input that is used to map concerns to bins.
Given a list of extracted concerns, supporting quotes, and binning guidance, our third LM-powered self-critique loop maps each concern to appropriate bins.
In practice, we process this mapping in batches, dynamically constructing the necessary response schema to guarantee an appropriate response (implemented via enumeration) for each concern (keyed by unique string prefix).

\paragraph{4. Summarize Bins}

While the intermediate outputs of the CommentNEPA pipeline can be useful to SMEs (\eg, for evaluation against ground-truth data; \Cref{sec:evals}), the terminal output of the application is a summary of the concerns in each bin, with citations to the original correspondence.


\section{Case-Study Evaluations}
\label{sec:evals}

We evaluate our framework and CommentNEPA on four case studies, each involving public comments received in response to projects requiring environmental review:
\begin{enumerate}
  \item (WS) Bureau of Land Management's Western Solar Plan \citep{WS}. 
  \item (CFFF) Columbia Fuel Fabrication Facility license renewal application \citep{CFFF}.
  \item (CPNP) Comanche Peak Nuclear Power Plant license renewal application \citep{CPNP}.
  \item (MBTA) Fish and Wildlife Service Migratory Bird Permits \citep{MBTA}.
\end{enumerate}
We first explain the manual workflow that SMEs performed, against which we evaluate our application, noting that there are important differences which preclude a direct ``apples-to-apples'' comparison.

For many projects that have similar scope, such as nuclear power plant license renewal applications, experienced SMEs anticipate the types of comments that the public may submit and the categories they may be separated into for a comment-response report.
SMEs may thus begin the comment review process with ``binning guidance'' that outlines these anticipated categories of concerns and explanations that assist with consistent disambiguation (\eg, examples of what types of comments belong or do not belong in each bin).
When reading letters submitted by the public, SMEs use the binning guidance to parse the text into substantive, in-scope resource- or concern-specific bins, while separating out comments that are out of scope or non-substantive into bins for more generalized responses.

At a high level, it is important to note the distinction between the SME workflow and the CommentNEPA pipeline (\Cref{sec:commentnepa}):
While CommentNEPA attempts to distill the key concerns of a letter, textually supported by the most relevant original quotes, SMEs focus extracting sections of text that might stand alone as substantive comments and them map each, separately, to a related bin.
In practice, we observe that the workflow adopted by SMEs leads to a tension between capturing the context of a comment (and thus select long, contiguous runs of text) acknowledging the multiple, simultaneous concerns that text often identifies (requiring the text be delimited into many separate pieces that can be independently mapped to bins).
This tension manifests in variations of style amongst SMEs, who identify each other as ``lumpers'' or ``splitters,'' accordingly.
CommentNEPA does not have this same deficiency, but, as we observe, the emphasis on extracting the independent concerns communicated by a letter of public correspondence substitutes generative text that paraphrases these concerns in place of runs of original text.

Despite the fact that CommentNEPA does not replicate the SME workflow directly, we still attempt to perform meaningful statistical comparisons with the data that we possess.
For each case study, we quantitatively compare the outputs of the application to the outputs of the manual work performed by SMEs.
In each case, the work performed by CommentNEPA is done \emph{without SME feedback} used to optimize prompts, but the target prompts in the underlying
self-critique loops are optimized subject to self-critiques.
By executing CommentNEPA in such an autonomous fashion, we establish a baseline evaluation for planned extensions of this work.
We perform this quantitative evaluation using two independent sets for each case study.

\begin{table}[h!]
\centering
\caption{Statistics comparing CommentNEPA, executed autonomously, to SME outputs across our four case studies.}
\label{tab:results}
\begin{tabular}{lcccccc}\toprule
Case Study & Unique Letters & Quote Recall & Quote Precision & Binning Recall & Binning Precision & Bins \\
\midrule
CFFF & 49   & 47.0\% & 90.7\% & 57.1\% & 53.8\% & 19  \\
CPNP & 82   & 30.1\% & 58.0\% & 42.0\% & 37.0\% & 39  \\
MBTA & 287  & 17.9\% & 80.8\% & 48.9\% & 49.8\% & 26  \\
WS   & 1198 & 23.5\% & 72.7\% & 50.7\% & 35.7\% & 62  \\
\midrule
All  & 1616 & 23.1\% & 73.7\% & 49.9\% & 38.1\% &     \\
\bottomrule
\end{tabular}
\end{table}

\begin{figure}[h!]
\centering
\includegraphics[width=4.5in,trim={0 0 0 2em},clip]{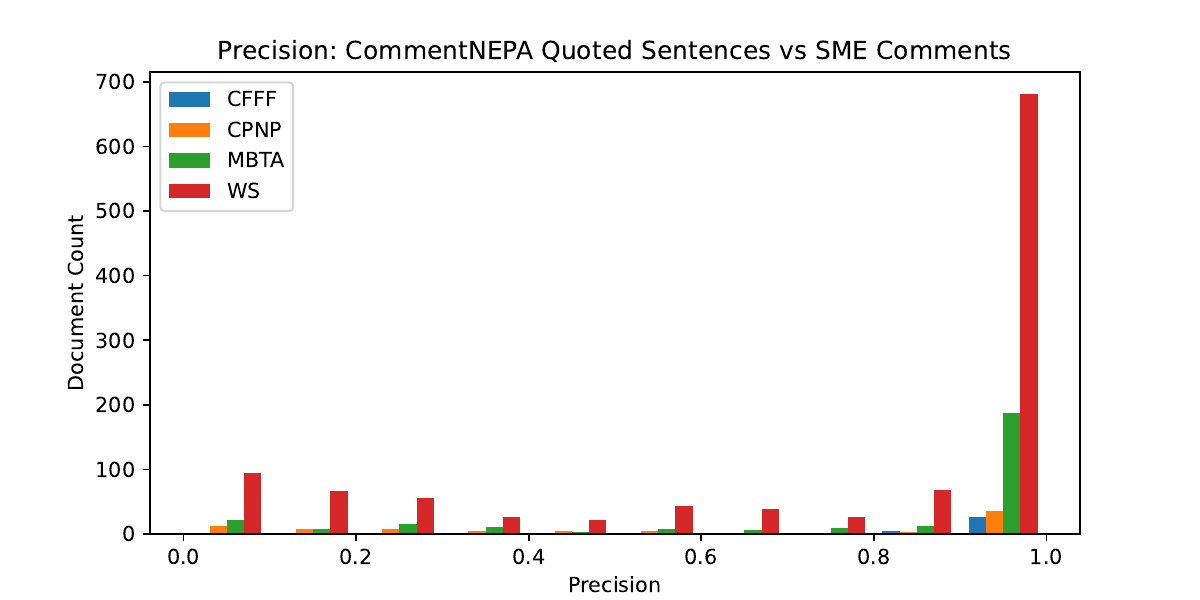}
\caption{The distribution of CommentNEPA's quote-selection precision across case studies. 
We visualize the number of documents whose sentences were selected with various precision ranges, where ``precision'' here is the probability that a given sentence of source correspondence selected by CommentNEPA to substantiate a claim was also selected by an SME-when delimiting excerpts of the correspondence to be independently binned  (\Cref{eq:precision}). 
We observe that this precision was consistently high across all four case studies.}
\label{fig:precision}
\end{figure}

\begin{figure}[h!]
\centering
\includegraphics[width=4.5in]{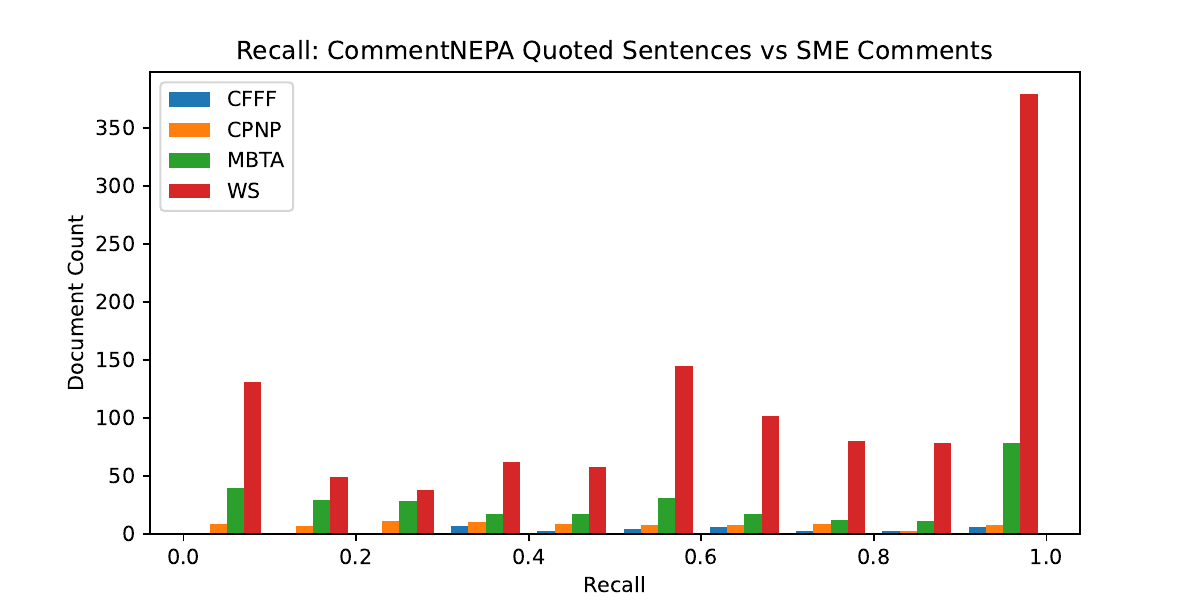}
\caption{The distribution of CommentNEPA's quote-selection recall across case studies. 
We visualize the number of documents whose sentences were selected with various recall ranges, where ``recall'' here is the probability that a given sentence of source correspondence selected by an SME-when delimiting excerpts of the correspondence to be independently binned was also selected by CommentNEPA to substantiate a claim (\Cref{eq:recall}). 
We observe that recall is not as consistently high as precision (\Cref{fig:precision}). Tellingly, recall is strongly dependent on the length of the source document (\Cref{fig:recall-vs-length}}.
\label{fig:recall}
\end{figure}

\paragraph{Comparing CommentNEPA's Quotes to SMEs' Substantive Comments}
For each letter of correspondence, we consider how the quotes selected by CommentNEPA compare to the ``substantive comments'' identified by SMEs in a manual workflow.
We quantify the performance on quote extraction by delimiting individual sentence in the source letter and treating the inclusion of each sentence in the set of selected quotes vs inclusion of the sentence in an SME-selected ``comment'' as a binary classification problem.
Upon exploring several figures of merit for such a binary classification task (e.g., Pearson correlation coefficient, accuracy, recall, precision, and F1 score), the salient pattern that emerged was that of consistently high precision and relatively low recall:
\begin{align}
\label{eq:precision}
\text{Precision:} \qquad &\Pr\Big(\text{Sentence} \in \text{SME-Comments} ~\Big|~ \text{Sentence} \in \text{Quotes} \Big) \\
\label{eq:recall}
\text{Recall:} \qquad &\Pr\Big(\text{Sentence} \in \text{Quotes} ~\Big|~ \text{Sentence} \in \text{SME-Comments}\Big)
\end{align}
These figures are reported per case study in the third and fourth columns of \Cref{tab:results} and further broken down into distributions for each case study in \Cref{fig:precision} and \Cref{fig:recall}.

\begin{figure}[h!]
\centering
\includegraphics[width=4.5in,trim={0 0 0 2em},clip]{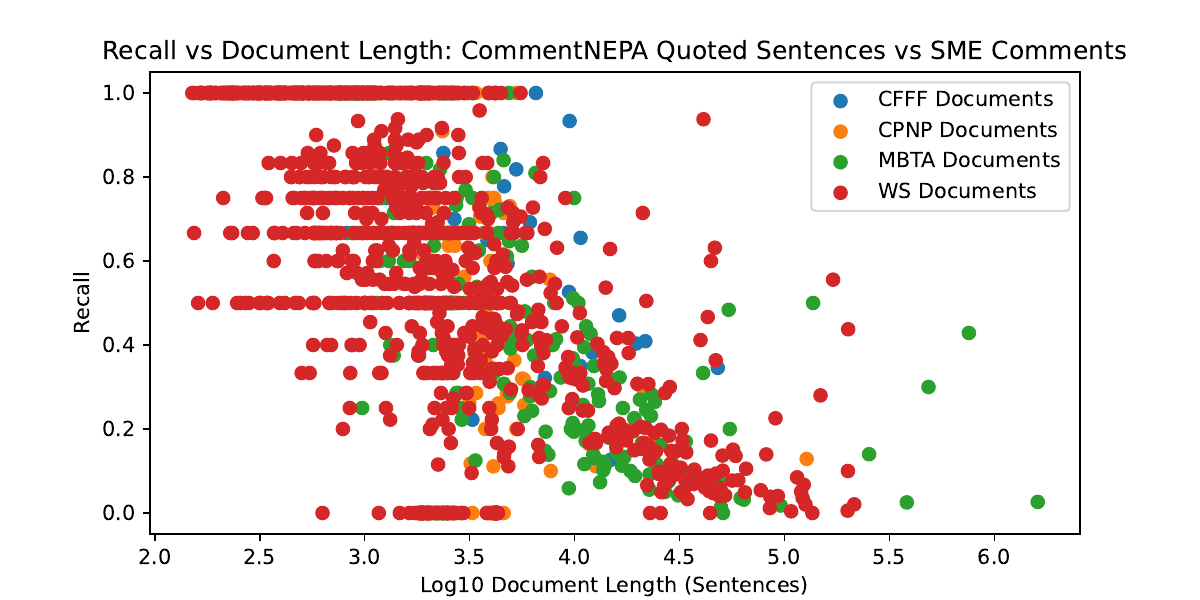}
\caption{The recall of sentences selected by SME (as independent, substantive comments) among set of quotes extracted by LM (substantiating extracted concerns) (\Cref{eq:recall}) as a function of document length. 
We observe that recall falls with document length.}
\label{fig:recall-vs-length}
\end{figure}

We visualize the precision (\Cref{eq:precision}) and recall (\Cref{eq:recall}) sentence-level statistics of CommentNEPA on quote extraction as measured against SME selections when delimiting independent text for binning. 
The consistently high precision (\Cref{fig:precision}) that we measure in this comparison suggests that CommentNEPA is likely extracting highly relevant quotes to support its extracted concerns: it vary rarely selects quotes that were not also considered substantive by SMEs.
Attending to recall (\Cref{fig:recall,fig:recall-vs-length}), our observation of declining recall with document length is consistent with at least four possible explanations:
\begin{enumerate}
    \item Contemporary LMs' ``needle-in-haystack'' recall declines over longer context windows, indicating that potentially relevant quotes are ignored during execution of CommentNEPA's concern extraction subroutine.
    \item When instructed to extract relevant quotes in support of an extracted concern, contemporary instruction-tuned LMs may be biased towards selecting relatively few quotes, thus ignoring others.
    \item Our task definition for quote extraction fundamentally differs from what SMEs do. In particular, we expect our task definition to yield only the most relevant quotes to each semantically distinct claim, rather than all quotes that an SME might select.
    \item LMs are more consistent, precise machines than SMEs, among whom we observe variance in opinions and selections. In particular, some SMEs (``lumpers,'' discussed earlier) tend to error on the side of caution, selecting long contiguous runs of text that will naturally suppress recall as measured here. 
    In support of this hypothesis, we refer to \Cref{tab:inconsistency}, which shows that across all projects, the length of comments identified by SMEs varied in length with a standard deviation of 120\%, while the length of quotes extracted by CommentNEPA varied in length by 55\%, demonstrating greater consistency in outputs.
\end{enumerate}

\begin{table}[h!]
\centering
\caption{Normalized standard deviation in length of relevant comments (quotes) extracted by SMEs (CommentNEPA).}
\label{tab:inconsistency}
\begin{tabular}{lcc}\toprule
case study & Norm. stdev. of comment length (SMEs) & Norm. stdev. of quote length (CommentNEPA) \\
\midrule
CFFF & 67\% & 52\% \\
CPNP & 116\% & 48\% \\
MBTA & 99\% & 54\% \\
WS & 122\% & 55\% \\
All & 120\% & 55\% \\
\bottomrule
\end{tabular}
\end{table}

\begin{figure}[h!]
\centering
\includegraphics[width=4.5in,trim={0 0 0 2em},clip]{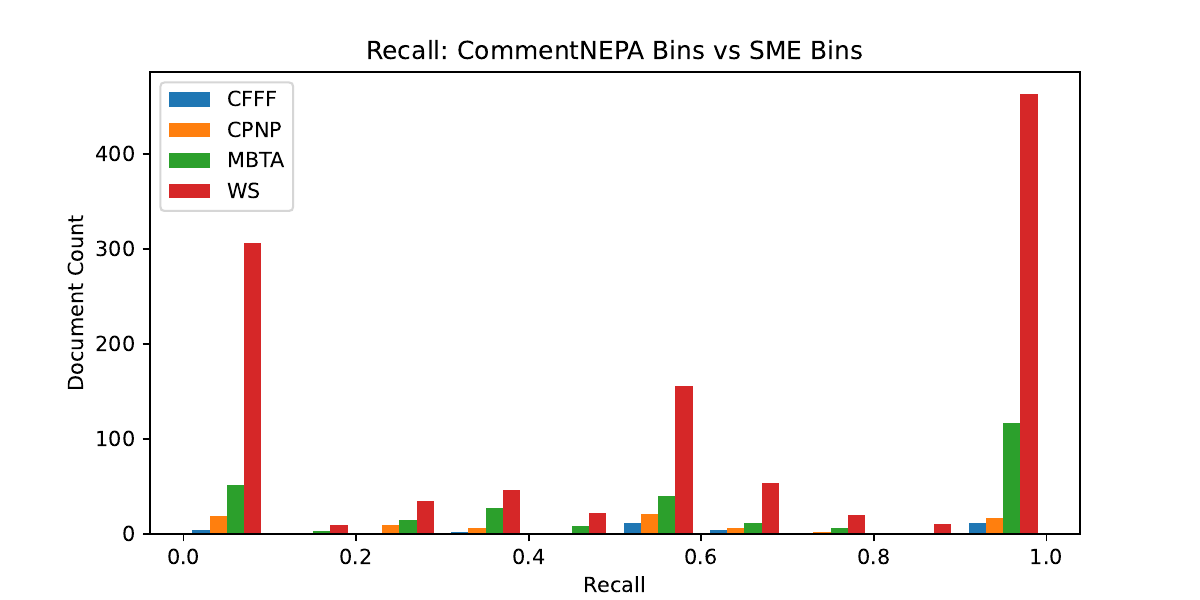}
\caption{We count the number of documents for which CommentNEPA achieved ranges of binning recall (\ie, for a given document and bin that an SME mapped it to, we consider the probability that CommentNEPA also mapped some of the letter's text to the same bin).}
\label{fig:bin-recall}
\end{figure}

\paragraph{Comparing Bins per Letter} 
In addition to sentence-selection statistics, for each letter of correspondence, we also compare how the set of ``bins'' (predefined categories of concerns) to which CommentNEPA mapped the text of a unit of public correspondence compares to the set of bins that an SME identified for the same text.
Once again, this is not an ``apples-to-apples'' comparison, because CommentNEPA is able to use the same sentence of source text in support of multiple extracted concerns that are individually mapped to bins;
The SMEs workflow that we compare to maps each unit of delimited text to only one bin.
We visualize the distribution of binning recall, or the probability that a bin identified by an SME was also identified by CommentNEPA per letter of correspondence, in \Cref{fig:bin-recall}.

As indicated in \Cref{tab:results}, binning recall was consistently \(\approx\)50\%  across all four case studies: Given a category of concerns identified by an SME for a randomly selected letter of correspondence, CommentNEPA identified the same category about half of the time.
To judge this metric, note that CommentNEPA was provided with binning guidance for each project, ranging from 19 bins with no instructions (CFFF) to 62 bins with detailed instructions (WS).
With such a large number of bins, 50\% recall and 40\% precision is indicative of accurate identification of categories that might be consistently confused with each other.
Indeed, \Cref{tab:results} shows that binning precision scales inversely with the number of identifiable bins.

\section{Conclusion and Future Work}

We have proposed a methodology for optimizing declarative, LM-powered subroutines within an $\infty$-MAB formalism, leveraging sparse human feedback and self-critique loops to improve the performance of these subroutines over time.
This methodology is implemented within an open-source software library on top of which we have developed and evaluated a first iteration of ``CommentNEPA,'' an application specialized for use by SMEs to accelerate NEPA's public commenting process.

Through our initial evaluations, we find that CommentNEPA is capable of extracting and sorting data with outputs that are comparable to those of subject matter experts, with potentially less variance and greater precision. 
Nonetheless, it remains to be determined whether critical information from the corpus of public correspondence is lost, especially in longer documents; whether SME feedback is able to improve the system over time as desired; and whether the potential for on-demand auditability can be presented in a way that minimizes undue burden for SMEs.

The development of CommentNEPA and refinement of the underlying techniques identified by our proposed methodology is ongoing, with plans to continue evaluations of the pipeline on additional case studies and to develop an online web interface with the ability to capture robust user feedback for continued online improvement.
Our goal is to realize the potential for language models to benefit the public through the acceleration of time-intensive tasks in governmental and public sector work while minimizing risk to the environment and legal processes; maintaining public trust; and establishing responsible paradigms for transparency, oversight, and audit.

\section{Acknowledgement}
This work was supported by the Office of Policy, U.S. Department of Energy, and Pacific Northwest National Laboratory, which is operated by Battelle Memorial Institute for the U.S. Department of Energy under Contract DE-AC05–76RLO1830.  This preprint has been cleared by PNNL for public release as PNNL-SA-213042.

\bibliographystyle{plainnat}
\bibliography{references.bib}

\newpage
\appendix
\section{Additional Figures}

\begin{figure}[tbh]
\begin{tcolorbox}[title=Example Case-Study Context (WS)]
To address changes since 2012, the Bureau of Land Management (BLM) is updating its to guidance on solar energy proposals. The BLM seeks to identify solar application areas with fewer resource conflicts, excluding areas with known sensitive resources, while also attempting to meet the national need for expanded solar energy development. This updated western solar plan encompasses 11 states, applies to solar projects 5 megawatts or larger, strengthens exclusion criteria since 2012, and removes development limits based on solar intensity. With the proposed changes, five alternatives have been identified, offering a range of different exclusion criteria and acreage available for solar development. In response to the BLM's proposals, the public has submitted correspondence that must be considered: The scope of issues and potential environmental consequences identified by the public must be comprehensively accounted for to move the proposal forward, appropriately adapt it, or reach any final determinations.
\end{tcolorbox}
\caption{The project description for the BLM's Western Solar (WS) Development Plan provided to relevant LM-powered subroutines in the CommentNEPA pipeline.}
\label{fig:ws-context}
\end{figure}

\begin{figure}[tbh]
\centering
\begin{tcolorbox}[title=Excerpt: A Typical Example of Sentences selected by SMEs vs CommentNEPA (WS), use color stack]
\sme{I fully support the BLMs efforts to make more public lands available for solar development but I cannot support the preferred alternative chosen by the BLM. }\both{Rather, I ask the BLM to choose alternative 5 with no provision for variances, or at least an extremely circumscribed provision for variances.}\sme{ In offering these comments I largely endorse the recommendations made by [redacted], in this recent post. [redacted] }\both{As [redacted] points, out the BLMs preferred Alternative 3 would open 22 million new acres of public lands to solar development based largely on the proximity of those lands to transmission lines.}\sme{ Two other alternatives would limit development to 11 million acres and 8.4 million acres respectively. The first of these, Alternative 4, would focus development on disturbed lands; the second, Alternative 5, would account for both disturbed lands and proximity to transmission lines. }\both{Alternative 5 is the most sensible option here.}\sme{ While we may need to use more of our public lands to promote renewable energy development, the BLM should most certainly not open public lands that are largely undisturbed, or that are distant from a power line. Doing so will invite litigation. Plainly, initial efforts should focus entirely on lands that are both disturbed and near a power line. }\both{Alternative 5 still makes 8.4 million acres of public land available and these lands should plainly receive priority for solar development, even if they are not necessarily the public lands most prized by developers.}

{\vspace{1em}\hfill {\bf Legend: } \sme{Used by SME only.} \llm{Used by LM only.} \both{Used by both.} Used by neither. \hfill}
\paragraph{SME Mapping:} {\bf Bin:} Preferred Alternative. {\bf Excerpt:} Entire text.
\paragraph{CommentNEPA Mapping:}
\begin{enumerate}
    \item {\bf Bin:} Preferred Alternative. {\bf Concern:} BLM's preferred alternative (Alternative 3) opens 22 million acres of public lands to solar development, potentially impacting undisturbed lands. {\bf Quotes:}
    \begin{itemize}
        \item ``As [redacted] points, out ...''
    \end{itemize}
    \item {\bf Bin:} Preferred Alternative. {\bf Concern:} Advocacy for Alternative 5, which prioritizes disturbed lands and proximity to transmission lines. {\bf Quotes:}
    \begin{itemize}
        \item ``Rather, I ask...''
        \item ``Alternative 5 is the most sensible...''
        \item ``Alternative 5 still makes...''
    \end{itemize}
\end{enumerate}
\end{tcolorbox}
\caption{A comparison of comment delineation and binning performed by SMEs and CommentNEPA on an excerpt of public correspondence received in response to the BLM's Western Solar (WS) proposal. 
This is a qualitatively typical example, truncated to exclude text that neither SME nor LM selected.}
\label{fig:sentences-representative}
\end{figure}

\begin{figure}[tbh]
\centering
\begin{tcolorbox}[title=Excerpt: An Example of Disagreement in Sentences selected by SMEs vs CommentNEPA (WS), use color stack]
\sme{Let's not keep making more mistakes that imperil the environment. In our quest
to transition to renewal energy,  
which I strongly support, we need to -- and can -- ensure that the development
of the necessary infrastructure does  
not create new harm.  
Producing sufficient energy is not, and never has been, an "either -- or"
choice between protecting the environment  
and having secure, stable, ample energy. Do not allow the industry lobbyists
to succeed in forcing you to make a  
choice that you do not need to make. For too long, we have allowed such
interests to dictate actions that have  
proven to have nothing short of disastrous consequences that will be very
expensive to remedy.} \llm{I support solar energy as a part of the United States' renewable energy
future. But it must be done in the right places  
to avoid harming sensitive biodiversity and cultural landscapes.}
\llm{As outlined in the Western Solar Plan draft environmental impact statement,
Alternative 5 would ensure that  
development is focused on lands that are no longer pristine and have already
been disturbed or have "low  
intactness." The Bureau of Land Management should choose this alternative,
which would keep most high-quality  
wildlife habitat off the table for energy development.}  
\llm{In addition, the agency needs to expand the exclusion zones -- areas excluded
from energy development no matter  
what. Expanded exclusion zones should include lakeshores, areas with sensitive
aquatic ecosystems or groundwater  
basins, remote areas where other development hasn't occurred, and occupied
endangered species habitats.}  
\llm{I believe solar energy is crucial but needs to be done in ways that don't harm
wildlife or rip up pristine public lands.  
Picking alternative 5 and expanding exclusion zones would avoid those harms,
while making more than enough land  
available for solar development.} 
To safely expedite renewable energy development, the BLM should choose
Alternative 5 and expand exclusion  
zones.

{\vspace{1em}\hfill {\bf Legend: } \sme{Used by SME only.} \llm{Used by LM only.} \both{Used by both. }Used by neither. \hfill}
\paragraph{SME Mapping:} {\bf Bin:} Preferred Alternative. {\bf Excerpt:} Let's not keep making...
\paragraph{CommentNEPA Mapping:}
\begin{enumerate}
    \item {\bf Bin:} General Impacts. {\bf Concern:} Potential harm to sensitive biodiversity and cultural landscapes from solar energy development. {\bf Quotes:}
    \begin{itemize}
        \item ``I support solar energy...''
    \end{itemize}
    \item {\bf Bin:} Preferred Alternative. {\bf Concern:} Advocacy for Alternative 5 to focus development on already disturbed lands. {\bf Quotes:}
    \begin{itemize}
        \item ``As outlined...''
    \end{itemize}
    \item {\bf Bin:} Resource-Based Exclusions. {\bf Concern:} Need for expanded exclusions zones. {\bf Quotes:}
    \begin{itemize}
        \item ``In addition, the agency needs...''
    \end{itemize}
    \item {\bf Bin:} Wildlife. {\bf Concern:} Avoiding harm to wildlife and pristine public lands. {\bf Quotes:}
    \begin{itemize}
        \item ``I believe solar energy..''
    \end{itemize}    
\end{enumerate}
\end{tcolorbox}
\caption{A comparison of comment delineation and binning performed by SMEs and CommentNEPA on an excerpt of public correspondence received in response to the BLM's Western Solar (WS) proposal. 
This example represents a minority of cases in which CommentNEPA's selections anti-correlate with the selections of an SME, though we argue that the appended binning information produced by CommentNEPA offers interpretability and a reasonable justification for its output.
We note the existence of artifacts of optical character recognition (the omission of spaces between words) that we aim to correct in future revisions.}
\label{fig:sentences-representative-disagreement}
\end{figure}

\begin{figure}[h!]
\centering
\includegraphics[width=5in]{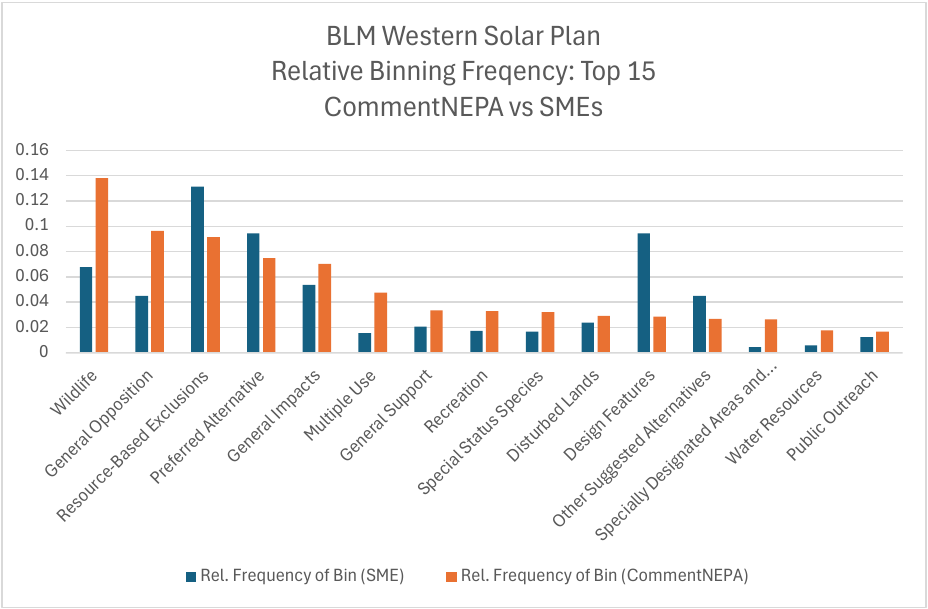}
\caption{The relative frequency distribution of bins (concern categories) identified in text by CommentNEPA and by SMEs for the BLM's Western Solar (WS) Development Plan. 
This selection represents 65\% of bin-mappings performed by SMEs and 75\% performed by CommentNEPA 
out of 62 total possible bins.}
\label{fig:bin-precision}
\end{figure}

\begin{figure}[h!]
\centering
\includegraphics[width=5in]{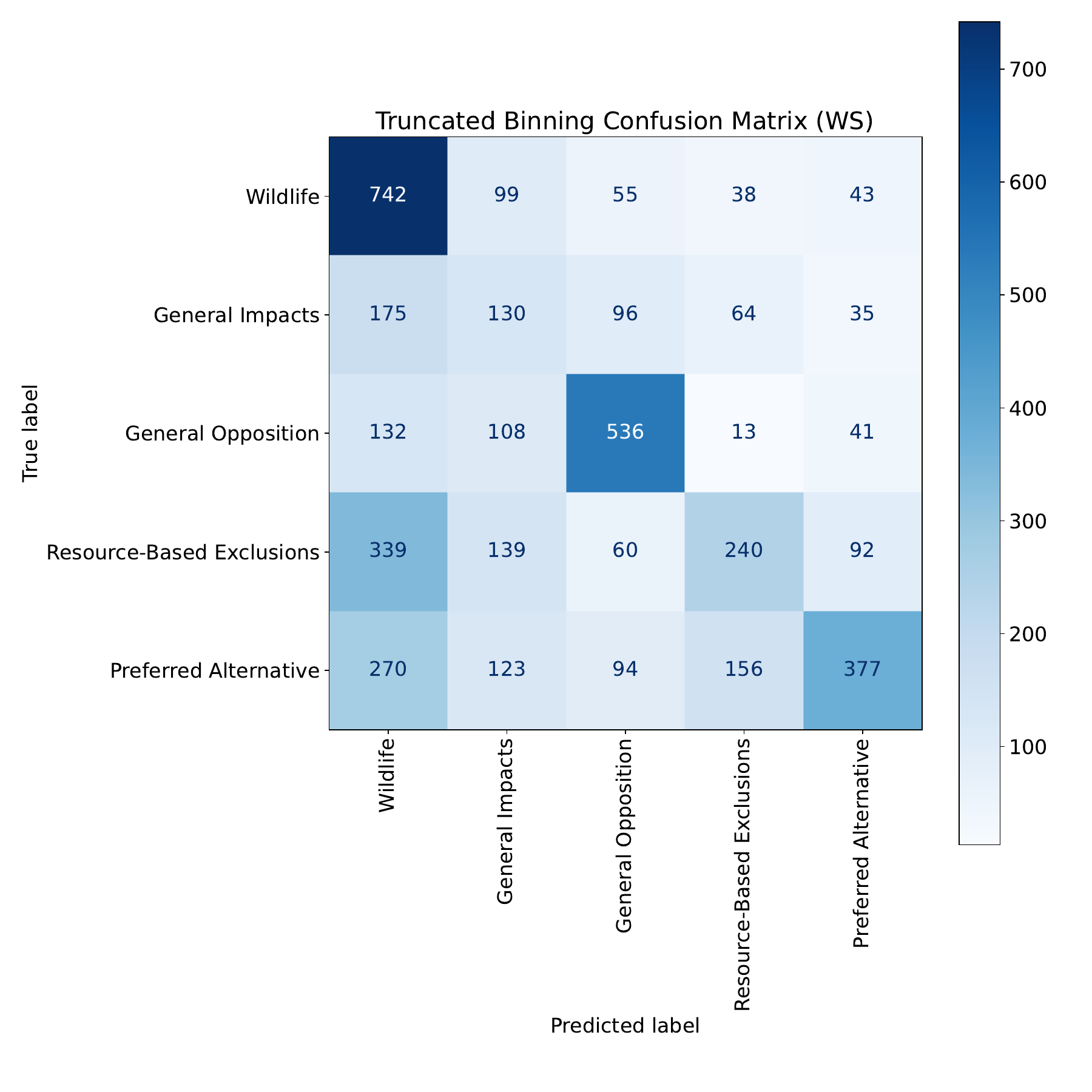}
\caption{A truncation of an (approximate) ``confusion matrix'' produced by CommentNEPA on the BLM's Western Solar (WS) Development Plan, retaining only the top 5 bins identified by SMEs.
To generate this ``confusion matrix'', we filter the corpus of correspondences for sentences that were both extracted by an SME and selected as a quotation by CommentNEPA: 
Numbers in this figure count the number of sentences that were associated with a (True label) bin by the SME and (Predicted label) bin by CommentNEPA, with the caveat that CommentNEPA has the capacity to map individual sentences to multiple bins when sentences are selected as quotes for multiple extracted concerns.
}
\label{fig:confusion-matrix}
\end{figure}

\end{document}